\documentclass[conference]{IEEEtran}
\IEEEoverridecommandlockouts
\usepackage{cite}
\usepackage{amsmath,amssymb,amsfonts}
\usepackage{algorithmic}
\usepackage{graphicx}
\usepackage{textcomp}
\usepackage{xcolor}
\usepackage[linesnumbered,ruled]{algorithm2e}

\usepackage{subcaption}

\def\BibTeX{{\rm B\kern-.05em{\sc i\kern-.025em b}\kern-.08em
    T\kern-.1667em\lower.7ex\hbox{E}\kern-.125emX}}
\begin{document}

\title{Differentiable Particle Filters through Conditional Normalizing Flow
\thanks{}
}

\author{\IEEEauthorblockN{Xiongjie Chen}
\IEEEauthorblockA{\textit{Department of Computer Science} \\
\textit{University of Surrey}\\
Guildford, UK\\
xiongjie.chen@surrey.ac.uk}
\and
\IEEEauthorblockN{Hao Wen}
\IEEEauthorblockA{\textit{Department of Computer Science} \\
\textit{University of Surrey}\\
Guildford, UK\\
h.wen@surrey.ac.uk}
\and
\IEEEauthorblockN{Yunpeng Li}
\IEEEauthorblockA{\textit{Department of Computer Science} \\
\textit{University of Surrey}\\
Guildford, UK \\
yunpeng.li@surrey.ac.uk}
}

\maketitle

\begin{abstract}
Differentiable particle filters provide a flexible mechanism to adaptively train dynamic and measurement models by learning from observed data. However, most existing differentiable particle filters are within the bootstrap particle filtering framework and fail to incorporate the information from latest observations to construct better proposals. In this paper, we utilize conditional normalizing flows to construct proposal distributions for differentiable particle filters, enriching the distribution families that the proposal distributions can represent. In addition, normalizing flows are incorporated in the construction of the dynamic model, resulting in a more expressive dynamic model. We demonstrate the performance of the proposed conditional normalizing flow-based differentiable particle filters in a visual tracking task.
\end{abstract}

\begin{IEEEkeywords}
Sequential Monte Carlo, Differentiable particle filters, Normalizing flows.
\end{IEEEkeywords}

\section{Introduction}
Particle filters, also known as sequential Monte Carlo (SMC) methods, are a family of algorithms designed for sequential state estimation tasks and have been applied in various domains including robotics~\cite{thrun2002particle}, computer vision~\cite{karkus2021differentiable}, target tracking~\cite{zhang2017multi}, and navigation~\cite{wang2017unbiased}. It requires the specification of the dynamic model which describes the transition of hidden state, and the measurement model which defines the likelihood of observation data given the predicted state. Existing parameter estimation methods for particle filters often assume that the structures and parts of dynamic and measurement models are known \cite{kantas2015particle}. However, it is non-trivial for practitioners to specify such models to simulate the true dynamics of states and their relation to observations, especially when dealing with complex environments in high-dimensional spaces. 

An emerging trend in developing data-adaptive particle filters is to learn the dynamic model and the measurement model through neural networks. Differentiable particle filters (DPFs) were proposed in \cite{jonschkowski-RSS-18, karkus2018particle} which provide a flexible way to learn the parameters of dynamic and measurement models. In \cite{ma2020particle}, DPFs were incorporated into recurrent neural networks (RNNs) to enable the modeling of multi-modal hidden state distribution. \cite{kloss2020train} proposed to learn noise terms in the dynamic and measurement models in DPFs adaptively from observation data. In \cite{corenflos2021differentiable}, a fully differentiable resampling method was proposed based on the entropy-regularized optimal transport.  A semi-supervised learning approach was introduced in \cite{wen2021end} to optimize the parameters of DPFs by iteratively maximizing a pseudo-likelihood function so that observations without ground truth state information can be used in the optimization of model parameters. These variants of DPFs \cite{jonschkowski-RSS-18, karkus2018particle,wen2021end} are within the bootstrap particle filtering (BPF) framework \cite{gordon1993novel}, in which particles are proposed according to the dynamic model. They hence do not utilize the latest observation in generating the proposal distribution and can potentially lead to the weight degeneracy issue.

Key requirements in constructing effective proposal distributions for DPFs include the expressiveness of the model in approximating the posterior distribution and the computational efficiency in evaluating the proposal densities. We consider normalizing flows ~\cite{de2020normalizing,abdelhamed2019noise,winkler2019learning} a promising probabilistic tool in constructing proposal distributions. Normalizing flows are a family of invertible neural networks that provide a general mechanism of constructing flexible probability distributions~\cite{papamakarios2021normalizing}. By transporting samples through an invertible neural network, normalizing flows can transform simple distributions (e.g. Gaussian distributions) into arbitrarily complex distributions under some mild conditions~\cite{papamakarios2021normalizing}, i.e. normalizing flows are universal approximators. Additionally, the evaluation of the densities of the transformed particles after applying normalizing flows is straightforward due to the invertibility of normalizing flows. One variant of normalizing flows, called conditional normalizing flow, can be  used  as  generative  models  conditioned  on  relevant data to generate conditional distributions~\cite{winkler2019learning}.

In this paper, we present a new variant of differentiable particle filters with flexible dynamic models and data-adaptive proposal distributions, which we call conditional normalizing flow-based differentiable particle filters. Our main contributions are three-fold: 1) We enrich distribution families that dynamic models can represent through the incorporation of normalizing flows; 2) We employ normalizing flows conditioned on latest observations to generate proposal distributions with efficiently evaluated proposal densities; 3) The proposed method can serve as a ``plug-in'' module in existing differentiable particle filter pipelines and we report the improved tracking performance in a visual tracking experiment.

The rest of the paper is organized as follows. Section~\ref{sec:ps} introduces the problem statement and Section~\ref{sec:background} provides the background information of the key components of the proposed work. We describe the proposed method in Section~\ref{sec:CNF_DPF} and report the experiment setup and results in Section~\ref{sec:experiment}. The conclusion is provided in Section~\ref{sec:conclusion}.

\section{Problem statement}
\label{sec:ps}

    We consider a nonlinear filtering task where the dynamic and measurement models are denoted as follows:
    \begin{align}
	    s_0&\sim \pi(s_0)\,\,, \\
	    s_t&\sim p(s_t| s_{t-1},a_t; \theta) \text{ for } t\geq1\,\,, \label{eq:transition} \\
        o_t&\sim p(o_t| s_t; \theta) \text{ for } t\geq1 \,\,\label{eq:likelihood}.
    \end{align}
	$\pi(s_0)$ is the stationary distribution of the hidden state at initial time $t=0$, $p(s_t| s_{t-1},a_t;\theta)$ is the dynamic model which describes the transition of hidden state $s_t$ at time step $t$ given the past state value $s_{t-1}$ and the action $a_{t}$ at the current time step. $p(o_t| s_t;\theta)$ is the measurement model which describes the relation between the observation $o_t$ and the hidden state $s_t$. $\theta$ denotes the set of parameters in the dynamic and measurement models. Our goal is to jointly learn the parameter set $\theta$ and estimate the marginal posterior distribution $p(s_t|o_{1:t},a_{1:t};\theta)$ or the joint posterior distribution $p(s_{1:t}|o_{1:t},a_{1:t};\theta)$, where $s_{1:t} = \{s_1, s_2, \ldots, s_t\}$, $o_{1:t} = \{o_1, o_2, \ldots, o_t\}$ and $a_{1:t} = \{a_1, a_2, \ldots, a_t\}$ are the history of states, observations and actions, respectively. 
	
\section{Background}
\label{sec:background}

\subsection{Particle filters}

    Particle filters approximate the posterior distribution by a set of weighted Monte Carlo samples~\cite{gordon1993novel}. The empirical estimate of the joint posterior distribution $p(s_{1:t}|o_{1:t}, a_{1:t}; \theta)$ can be expressed by:
    \begin{equation}
        {p}(s_{1:t}|o_{1:t}, a_{1:t}; \theta)\approx \sum_{i=1}^{N_p}w_t^i \delta(s_{1:t}-s_{1:t}^i)\,\,,
    \end{equation}
    where $s_{1:t}^i$ are the history of state values for the $i$-th particle, $\delta(\cdot)$ is the Dirac delta function, $w_t^i$ is the normalized weight of the $i$-th particle at time-step $t$. Particles  $s^i_{1:t}$ are sampled according to the proposal distribution $q(s_{1:t}|o_{1:t}, a_{1:t}; \theta)$ where with a slight abuse of notation, we use $\theta$ to also denote the parameters of the proposal distribution. The importance weight associated with the $i$-th particle is given by:
    \begin{equation}\label{eq:importance_weight}
        w_t^i \propto \frac{p(s^i_{1:t}|o_{1:t}, a_{1:t};\theta)}{q(s^i_{1:t}|o_{1:t}, a_{1:t}; \theta)}\,\,.
    \end{equation}
    Given the hidden Markov models described in Equations~\eqref{eq:transition}  and~\eqref{eq:likelihood},
    the proposal density can be factorized as:
    \begin{align}
        &q(s_{1:t}|o_{1:t}, a_{1:t};\theta)\nonumber\\
        &= q(s_t|s_{t-1}, o_{t}, a_{t};\theta) q(s_{1:t-1}|o_{1:t-1}, a_{1:t-1}; \theta)\,\,,
    \end{align}
    and the posterior distribution $p(s_{1:t}|o_{1:t}, a_{1:t};\theta)$ can be re-written as:
    \begin{align}
        &p(s_{1:t}|o_{1:t}, a_{1:t};\theta) \nonumber\\
        &= \frac{p(o_{t}|s_{t};\theta)p(s_t|s_{t-1}, a_t; \theta)}{p(o_t|o_{1:t-1}; \theta)}p(s_{1:t-1}|o_{1:t-1}, a_{1:t-1}; \theta)\nonumber\\
    &\propto p(o_{t}|s_{t};\theta) p(s_t|s_{t-1}, a_t; \theta) p(s_{1:t-1}|o_{1:t-1}, a_{1:t-1}; \theta)\,\,.
    \end{align}
    The importance weights can be updated as follows:
    \begin{align}\label{eq:recursive_weight}
    w_t^i &\propto \frac{p(s^i_{1:t}|o_{1:t}, a_{1:t};\theta)}{q(s^i_{1:t}|o_{1:t}, a_{1:t}; \theta)}\nonumber\\
    &\propto w_{t-1}^i \frac{p(o_{t}|s_{t}^i;\theta) p(s_t^i|s_{t-1}^i, a_t; \theta)}{q(s_t^i|s_{t-1}^i, o_{t}, a_{t};\theta)}\,\,.
    \end{align}
    
    A generic particle filtering algorithm is provided in Algorithm~\ref{SMC}.
    
    \begin{algorithm}[h]
	\begin{algorithmic}[1]
		\caption{A generic particle filtering algorithm}\label{SMC}
		\STATE \textbf{Require}: stationary distribution of state at initial time $\pi(s_0)$, resampling threshold $N_{thres}$, particle number $N_p$;
		\STATE \textbf{Initialisation}: Draw sample $\{s_0^i\}_{i=1}^{N_p}$ from $\pi(s_0)$;
		\STATE Set $\{w_0^i\}_{i=1}^{N_p}=\frac{1}{N_p}$;		
		\FOR {$t=1$ to $T$}
		\STATE Sample $\{s_t^i\}_{i=1}^{N_p}\sim q(s^i_t|s_{t-1}^i, o_t, a_t;\theta)$;
		\STATE $w_t^i=w_{t-1}^i \frac{p(o_{t}|s_{t}^i; \theta) p(s_t^i|s_{t-1}^i, a_t; \theta)}{q(s_t^i|s_{t-1}^i, o_{t}, a_{t}; \theta)}$;
		\STATE Normalize weights $\{w_t^i\}_{i=1}^{N_p}$ so that $\sum_{i=1}^{N_p}w_{t}^i=1$;
		\STATE Compute the effective sample size: $ESS_k=\frac{1}{\sum_{i=1}^{N_p}(w_t^i)^2}$;
		\IF {$ESS_t < N_{thres}$} 
		\STATE  Resample $\{s_t^i,w_t^i\}_{i=1}^{N_p}$ to obtain $\{s_t^i,\tfrac{1}{N_p}\}_{i=1}^{N_p}$;
		\ENDIF
		\ENDFOR			
	\end{algorithmic}
\end{algorithm}

    The choice of the proposal distribution influences the variance of particle filter estimations \cite{doucet2001introduction}. The ``optimal" proposal distribution $p (s_t| s_{t-1}, o_t, a_t; \theta)$ minimizes the variance of particle weights yet is rarely accessible in practical scenarios \cite{doucet2000sequential}.
    
    

\subsection{Differentiable particle filters}
    \label{subsec:DPFs}
    Differentiable particle filters use the expressiveness of neural network to model the dynamic and measurement models in particle filters. For the dynamic model, the relative motion of hidden states between time steps is typically modelled using a neural network \cite{jonschkowski-RSS-18, kloss2020train}:
    \begin{equation}
	\label{eq:dynamics_semi}
	    s_t^i=s_{t-1}^i+f_\theta(s_{t-1}^i, a_t)+\epsilon^i\sim p(s_t| s_{t-1}^i,a_t; \theta)\,\,,
	\end{equation}
	where $f_\theta$ is a neural network used to model the relative motion, and $\epsilon^i$ is an auxiliary noise vector with independent marginal distribution $p(\epsilon)$ used in the reparameterization trick \cite{kingma2013auto}.
	For the measurement model, the likelihood of the observation given the hidden state is calculated by $p(o_t| s_t; \theta)= l_\theta(o_t, {s}_t^i)$,
	where $l_\theta$ is a neural network, or a transformation of the neural network output, used to estimate the likelihood~\cite{jonschkowski-RSS-18,wen2021end}.
	
	The optimization of DPFs \cite{jonschkowski-RSS-18, karkus2018particle, ma2020particle, kloss2020train} relies on the ground truth state information, and the objective function can be the root mean square error (RMSE) between the ground truth state and the particle mean, or the negative log-likelihood (NLL) of the ground truth state under the approximated posterior distribution. In the semi-supervised DPFs (SDPFs)~\cite{wen2021end}, a pseudo-likelihood function is maximized to utilize unlabeled state in parameter optimization. Specifically, by dividing observations, actions, and states into $B$ blocks of length $L$, the pseudo-likelihood used in \cite{wen2021end} can be formulated as:
	\begin{align}
	    Q(\theta)&= \frac{1}{B}\sum_{b=0}^{B-1} \hat{Q}(\theta,\theta_{b})\,\,,
	\end{align}
	where $\hat{Q}(\theta,\theta_{b})$ is an estimation of the pseudo-likelihood at the $b$-th block. $\hat{Q}(\theta,\theta_{b})$ can be calculated as:
	\begin{align}
	    \hat{Q}(\theta,\theta_{b})=&\sum_{i=1}^{N_p}w_{(b+1)L}^i\log(\pi({s}^i_{bL+1}) p(o_{bL+1}| s^i_{bL+1}; \theta)\nonumber\\
        &\prod_{m=bL+2}^{(b+1)L}p({s}_m^i| s_{m-1}^i,a_m; \theta) p(o_{m} | {s}^i_{m}; \theta))\,,
	\end{align}
	where $w_{(b+1)L}^i$ is the particle weight of the $i$-th particle at time step $(b+1)L$, $\pi(s^i_{bL+1})$ is the stationary distribution of the $i$-th particle at time step $bL+1$, and $p({s}_m^i| s_{m-1}^i,a_m; \theta)$ is the density of proposed particles evaluated with the dynamic model. 

\subsection{Normalizing flows}
\label{subsec:normalizing_flows}
Normalizing flows are a family of invertible transformations that provide a general mechanism for constructing flexible probability distributions~\cite{papamakarios2021normalizing}. Let $u\in \mathbb{R}^d$ be a $d$-dimensional variable, and suppose $u$ follows a simple distribution $p(u)$. We can express a variable $x\in\mathbb{R}^d$ that follows a complex distribution $p(x)$ as a transformation $\mathcal{T}_\theta(\cdot)$ of variable $u$ following a simple distribution $p(u)$:
\begin{equation}
    x=\mathcal{T}_\theta(u)\sim p(x)\,,
\end{equation}
where $\theta$ is the parameter of the transformation, $u\sim p(u)$, and such a transformation $\mathcal{T}_\theta(\cdot)$ is called a normalizing flow if $\mathcal{T}_\theta(\cdot)$ is invertible and differentiable. 

With the above properties, we can evaluate the probability density of $p(x)$ by applying the change of variable formula:
\begin{equation}
    p(x)=p(u)|\text{det}J_{\mathcal{T}_\theta}(u)|^{-1}\,,
\end{equation}
where $|\text{det}J_{\mathcal{T}_\theta}(u)|$ is the Jacobian determinant of $\mathcal{T}_\theta(\cdot)$ evaluated at $u$.

One variant of normalizing flows, the Real NVP model~\cite{dinh2016density}, constructs the normalizing flow through the coupling layer. Denote by $u\in \mathbb{R}^d$ the input of a coupling layer, in the standard coupling layer the input $u$ is split into two parts $u=[u_1,u_2]$, where $u_1=\underset{1:d'}{u}$ refers to the first $d'$ dimensions of $u$, and $u_2=\underset{d'+1:d}{u}$ refers to the last $d-d'$ dimensions of $u$. The partition is uniquely determined by an index $d'<d$, and the output $x\in\mathbb{R}^d$ of the coupling layer is given by:
\begin{gather}
	\underset{1:d'}{x}=\underset{1:d'}{u}\,,\\
	\underset{d'+1:d}{x}=\underset{d'+1:d}{u}\odot \exp (c(\underset{1:d'}{u}))+t(\underset{1:d'}{u})\,,
\end{gather}
where $c: \mathbb{R}^{d'}\rightarrow \mathbb{R}^{d-d'}$ and $t: \mathbb{R}^{d'}\rightarrow \mathbb{R}^{d-d'}$ stand for the scale and translation function, $\odot$ is the element-wise product. By consecutively applying the above process to the outputs of coupling layers, the Real NVP model can construct arbitrarily complex probability distributions~\cite{papamakarios2021normalizing}.

In applications such as image super-resolution, normalizing flows are used as generative models conditioned on some relevant data such as low-resolution images, and this variant of normalizing flows is called conditional normalizing flows~\cite{winkler2019learning}. In order for the normalizing flow to be conditional, the standard coupling layers in the Real NVP are replaced by conditional coupling layers. In contrast to unconditional coupling layer, the scale and the translation function in conditional coupling layers are functions of concatenations of the original input $u_1$ and some relevant variable $z\in \mathbb{R}^D$, i.e. the input of the translation and the scale functions now become $[\underset{1:d'}{u}, z]$. Therefore, a conditional coupling layer is defined by:
\begin{gather}
	\underset{1:d'}{x}=\underset{1:d'}{u}\,,\\
	\underset{d'+1:d}{x}=\underset{d'+1:d}{u}\odot \exp (c(\underset{1:d'}{u},z))+t(\underset{1:d'}{u},z)\,,
	\label{eq:cnf}
\end{gather}
where $c:\mathbb{R}^{D+d'}\rightarrow\mathbb{R}^{d-d'}$ and $t:\mathbb{R}^{D+d'}\rightarrow\mathbb{R}^{d-d'}$ are functions defined on $\mathbb{R}^{D+d'}$.

\section{Conditional normalizing flow-based differentiable particle filters}
\label{sec:CNF_DPF}

In this section we provide details on the proposed algorithms, i.e. conditional normalizing flow DPFs (CNF-DPFs). Specifically, we show that conditional normalizing flows can be used to migrate particles generated by dynamic models to construct better proposals with tractable proposal densities. We also show that normalizing flows can be straightforwardly incorporated into dynamic models. We provide numerical implementation details for the proposed method based on the semi-supervised DPFs proposed in~\cite{wen2021end}, which we name as the CNF-SDPF. To simplify the notation, we use $\theta$ to denote the set of parameters in the dynamic model, the measurement model, and the proposal distribution in this section. 

\subsection{Dynamic model with normalizing flows}
\label{subsec:dyn_model_nfs}
We first show a direct approach to use normalizing flows to construct flexible dynamic models. We can first construct a ``prototype'' dynamic model $g(\tilde{s}_t|s_{t-1}^i,a_t; \theta)$ following simple distributions, e.g. Gaussian, obtained from prior knowledge of the physical settings that can coarsely predict particles in next steps. Then we use $\mathcal{T}_\theta(\cdot): \mathbb{R}^d \rightarrow \mathbb{R}^d$ to denote a normalizing flow parameterized by $\theta$ defined on a $d$-dimensional space. To enable dynamic models to capture complex distributions, the particles $\tilde{s}_t^i$ generated by $g(\tilde{s}_t|s_{t-1}^i,a_t; \theta)$ are further transported by the normalizing flow:

\begin{equation}
\hat{s}_t^i=\mathcal{T}_\theta(\tilde{s}_t^i)\,\,.
\label{eq:nf_dynamic}
\end{equation}

\subsection{Normalizing flows conditioned on observations} 
\label{subsec:cond_nf}

As in Equation (\ref{eq:cnf}), the input of conditional normalizing flows consists of two parts, the first is the original input and the second is a relevant variable. In the framework of DPFs, the original input is the particles generated by dynamic models and the relevant variable is the latest observation $o_t$, thus the conditional normalizing flow used in DPFs can be formulated as:
\begin{equation}
    s_t^i=\mathcal{G}_\theta(\hat{s}_t^i,o_t)\,\,,
\label{eq:cnf_proposal}
\end{equation}    
where $\hat{s}_t^i$ refers to the $i$-th particle propagated by the dynamic model given by Equation~(\ref{eq:nf_dynamic}), i.e. generated by first propagating the particle through $g(\cdot; \theta)$ then the unconditional normalizing flow $\mathcal{T}_\theta(\cdot)$. The conditional normalizing flow $\mathcal{G}_\theta(\cdot)$ is then applied on $\hat{s}_t^i$. For brevity and a slight abuse of notation, we denote the Jacobian determinant of the conditional normalizing flow by $\text{det}\,\, J_{\mathcal{G}_\theta}(\hat{s}_t^i)$ by omitting $o_t$ in the notation. The proposal density can be obtained by applying the change of variable formula:
\begin{align}
    q&(s_t^i|s_{t-1}^i,a_t,o_t; \theta)=p(\hat{s}_t^i|s_{t-1}^{i},a_t; \theta)\bigg|\text{det}\,\, J_{\mathcal{G}_\theta}(\hat{s}_t^i)\bigg|^{-1}\nonumber\\
    &=g(\tilde{s}_t^i| s_{t-1}^i,a_t; \theta) \bigg|\text{det}\,\, J_{\mathcal{T}_\theta}(\tilde{s}_t^i)\bigg|^{-1} \bigg|\text{det}\,\, J_{\mathcal{G}_\theta}(\hat{s}_t^i)\bigg|^{-1}\,,
    \label{eq:proposal_density}
\end{align}
where $s_t^i$, $\hat{s}_t^i=\mathcal{G}_\theta^{-1}(s_t^i, o_t)$, and $\tilde{s}_t^i=\mathcal{T}_\theta^{-1}(\mathcal{G}_\theta^{-1}(s_t^i, o_t))$ are the $i$-th particle generated by the proposal distribution (Equation~\eqref{eq:cnf_proposal}), the $i$-th particle generated by the dynamic model (Equation~\eqref{eq:nf_dynamic}), and the $i$-th particle generated by the ``prototype'' dynamic model $g(\cdot; \theta)$, respectively. In addition, the density of proposed particles evaluated with the dynamic model is:
\begin{equation}
    p(s_t^i|s_{t-1}^i,a_t; \theta)=g\big(\mathcal{T}_\theta^{-1}(s_t^i)|s_{t-1}^i,a_t; \theta\big)\big|\text{det}J_{\mathcal{T}_\theta}(\mathcal{T}_\theta^{-1}(s_t^i))\big|^{-1}\,.
    \label{eq:dynamic_density}
\end{equation}
Combining Equations~\eqref{eq:recursive_weight},~\eqref{eq:proposal_density} and~\eqref{eq:dynamic_density}, we can obtain the importance weight of the $i$-th particle as follows:
\begin{align}
    &w_t^i\propto\frac{p(s_t^i|s_{t-1}^i,a_t;\theta)p(o_t|s_t^i; \theta)w_{t-1}^i}{q(s_t^i|s_{t-1}^i,a_t, o_t; \theta)}\nonumber\\
    &=w_{t-1}^i\frac{g\big(\mathcal{T}_\theta^{-1}(s_t^i)|s_{t-1}^i,a_t; \theta\big)p(o_t|s_t^i; \theta)\bigg|\text{det} J_{\mathcal{T}_\theta}(\tilde{s}_t^i)\;\text{det} J_{\mathcal{G}_\theta}(\hat{s}_t^i)\bigg| }{g(\tilde{s}_t^i| s_{t-1}^i,a_t; \theta)  \big|\text{det}J_{\mathcal{T}_\theta}(\mathcal{T}_\theta^{-1}(s_t^i))\big| }\,.
    \label{eq:weights_update}
\end{align}


\subsection{Numerical implementation} 
The proposed conditional normalizing flow-based differentiable particle filter framework can be directly incorporated to different DPF variants. For illustration, we provide numerical implementation details of the framework based on the semi-supervised differentiable particle filter (SDPF) \cite{wen2021end} in Algorithm~\ref{alg:alg1}.


\begin{algorithm}[htpb]
	\caption{CNF-SDPF}
	\label{alg:alg1}
	\SetEndCharOfAlgoLine{}
	\SetKwComment{Comment}{// }{}
	\SetKwInOut{Input}{Input}
	\SetKwFor{For}{for}{do}{end~for}
	\Input{\\\hspace{-3.6em}\small
		\begin{tabular}[t]{l @{\hspace{.5em}} l}%
			$o_{1:T}$ & Observations \\
			$a_{1:T}$ & Actions \\
			$N_p$ & Particle number\\
			$\alpha$ & Learning rate \\
			$\lambda_1, \lambda_2$ & Scaling factors\\
			$g(\cdot ; \theta)$ & Prototype \\
			$l_\theta$ & Meas. model \\
			$\mathcal{D}$ & Cosine distance \\
			$h_\theta$ & Obs. encoder \\
			$\hat{h}_\theta$ & State encoder \\
		\end{tabular}\hspace{-0.5em}%
		\begin{tabular}[t]{l @{\hspace{.5em}} l}%
			$\pi(s_0)$ & Initial distribution of $s_0$\\
			$T$ & Episode time\\
			$s_t^*$ & Ground truth state\\
			$N_{thres}$ & Resampling threshold\\
			$L$ & Block length\\
			$f_\theta$ & Action transformer \\
			$\mathcal{S}(\cdot)$ & Supervised loss function\\
			$\epsilon$ & Reparam. noise vector\\
			$p(\epsilon)$ & Distribution of $\epsilon$\\
			$\beta$ & Trade-off parameter\\
		\end{tabular}%
	}
	\BlankLine
	Initialize parameters $\theta$ of $f_\theta$, $h_\theta$ and $\hat{h}_\theta$ randomly; 
	Draw particles $\{s_0^i\}_{i=1}^{N_p}$ from $\pi(s_0)$; 
	Set particle weights  $\{w_0^i\}_{i=1}^{N_p}=\frac{1}{N_p}$; 
	Set $Q=0$ and $b=0$;
	
	\While{$\theta$ not converged}{
		\For{$t=1$ to $T$ }{
			Compute the effective sample size: $N_{eff}=\frac{1}{\sum_{i=1}^{N_p}(w_{t-1}^i)^2}$;
			
			\uIf{$N_{eff}<N_{thres}$}{
			    $\{v_{t-1}^i\}_{i=1}^{N_p}=\{\beta w_{t-1}^i+(1-\beta)\frac{1}{N_p}\}_{i=1}^{N_p}$;
			    
				Select ancestor index $A_{t-1}^i$ with $Pr(A_{t-1}^i=j)=v_{t-1}^j$ for $i,j=1,...,N_p$;
				
				Update particle weights: $\{w_{t-1}^i\}_{i=1}^{N_p}=\{\frac{w_{t-1}^i}{v_{t-1}^i}\}_{i=1}^{N_p}$;
				
				Normalize weights $\{w_{t-1}^i\}_{i=1}^{N_p}$;
			}
			\Else{
				$A_{t-1}^i=i$ for $i=1,...,N_p$;
			}
			Set $s_{0:t-1}^i=s_{0:t-1}^{A_{t-1}^i}$ for $i=1, ..., N_p$;
			
			$e_t=h_\theta(o_t)$; 
			
			\For{$i=1$,...,$N_p$}{
				Draw particles $\tilde{s}_t^i \sim g(\tilde{s}_t|s_{t-1}^i,a_t; \theta)=s_{t-1}^i+f_\theta(s_{t-1}^i, a_t)+\epsilon^i$ where $\epsilon^i\sim p(\epsilon)$;\newline
				Transform particles with normalizing flows: $\hat{s}_t^i=\mathcal{T}_\theta(\tilde{s}_t^i)$;
				
				Migrate particles to proposal distributions with conditional normalizing flows: $s_t^i=\mathcal{G}_\theta(\hat{s}_t^i, e_t)$;

				Calculate $q(s_t^i|s_{t-1}^i,a_t,e_t; \theta)$ and $p(s_t^i|s_{t-1}^i,a_t; \theta)$ using Equation~(\ref{eq:proposal_density}) and Equation~(\ref{eq:dynamic_density}), respectively;
				
				$\tilde{e}^i_t=\hat{h}_\theta(s_t^i)$, $l_\theta(o_t, s_t^i)=1/\mathcal{D}(\tilde{e}^i_t, e_t)$;
				
				Update particle weights: $w_t^i=w_{t-1}^i  \frac{p(s_t^i|s_{t-1}^i,a_t;\theta)l_\theta(o_t, s_t^i)}{q(s_t^i|s_{t-1}^i,a_t, e_t; \theta)}$;
			}
			Normalize weights $\{w_t^i\}_{i=1}^{N_p}$ so that $\sum_{i=1}^{N_p}w_{t}^i=1$;
			\uIf{$t \mod L=0$}{
				\For{$i=1,...,N_p$}{
					$\eta^i=\pi(s^i_{bL+1})l_\theta(o_{bL+1}, s^i_{bL+1})$
					$\,\,\,\,\,\,\prod_{m=bL+2}^{(b+1)L}p(s_m^i|s_{m-1}^i,a_m; \theta) l_\theta(o_{m},{s}^i_{m})$;
				}
				Compute $Q=Q+\sum_{i=1}^{N_p}w_{t}^i\log(\eta^i)$;
				$b=b+1$;
			}
		}
		Total loss: $\mathcal{L}=\lambda_1 \frac{1}{|T|}\sum_{t=1}^{T} \mathcal{S}(s_t^*, \{w_t^i, s_t^i\}_{i=1}^{N_p})-\lambda_2 \frac{1}{b}Q$ ;
		Update model parameters $\theta\leftarrow\theta-\alpha\nabla_\theta \mathcal{L}$; 
	}
\end{algorithm}

	               





\section{Experiment results}
\label{sec:experiment}

We study a synthetic visual tracking environment that has been used to evaluate differentiable filters \cite{haarnoja2016backprop, kloss2020train, yi2021differentiable}. Here, we compare performance of the differentiable particle filter (DPF)~\cite{jonschkowski-RSS-18}, semi-supervised differentiable particle filter (SDPF)~\cite{wen2021end}, and the proposed conditional normalizing flow-based differentiable particle filters based on these two DPF variants, named as the CNF-DPF and the CNF-SDPF, respectively. 

\subsection{Problem setup}

\begin{figure}[th]
	\begin{center}
		\includegraphics[width=0.42\linewidth]{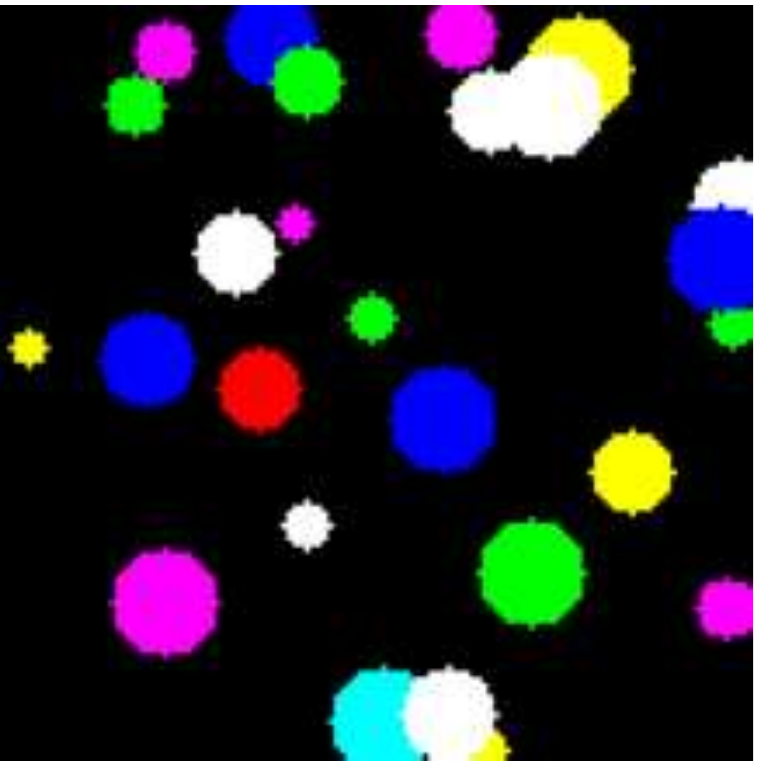}
		\caption{Observation images in the disk tracking experiment, the aim is to track the red disk among distractors.}
		\label{fig:red_disk}
	\end{center}
\end{figure}

 The objective for the simulated disk tracking is to track a red disk moving along the other distractor disks as shown in Fig. \ref{fig:red_disk}. The number of distractors can be changed accordingly to create tasks with different levels of difficulty. The colour and the size of the distractors are generated randomly - the colours of the distractors do not include red (the target) and black (the background). The red disk can be occluded by the distractors. The disks can temporally leave the frame since the contacts to the boundary are not modelled.

The observation images are $128\times128\times3$ RGB images. The positions of the red disk and distractors are uniformly distributed in the observation image, and the velocity of the red disk and distractors are sampled from the standard normal distribution. The radius of the red disk is set to be $7$, and the radii of the distractors are randomly sampled with replacement from $\{3,4,...,10\}$. The number of distractors is set to be 25.

The hidden state at current time $t$ is denoted as $s_t$, which represents the position of the target red disk. The action at current time step $t$ is denoted as $a_t$, which is the velocity of the target. Following the setup in \cite{kloss2020train}, the state evolves as follows:

\begin{align}
\hat{a}_t &= a_t + \epsilon_a\,\,,\\ 
s_{t+1} &= s_t + \hat{a}_t + \epsilon\,\,,
\label{eq:dynamic1}
\end{align}
where $\hat{a}_t$ is the noisy action from the given input $a_t$ corrupted by multivariate Gaussian noises $\epsilon_a\sim \mathcal{N}(0, 4^2)$, and $\epsilon\sim\mathcal{N}(0, 2^2)$ is the dynamic noise.

Each trajectory has $50$ time steps, and the mini-batch size is set to be $32$. The dataset contains $400$ trajectories for training, $50$ trajectories for validation and $50$ for testing. The number of particles used for both training and testing is set to be $N_p=100$. Particles are initialized around the true state with noises sampled from standard normal distributions.

\subsection{Experiment results}

\begin{figure}[t]
    \begin{center}
    \includegraphics[width=0.95\linewidth]{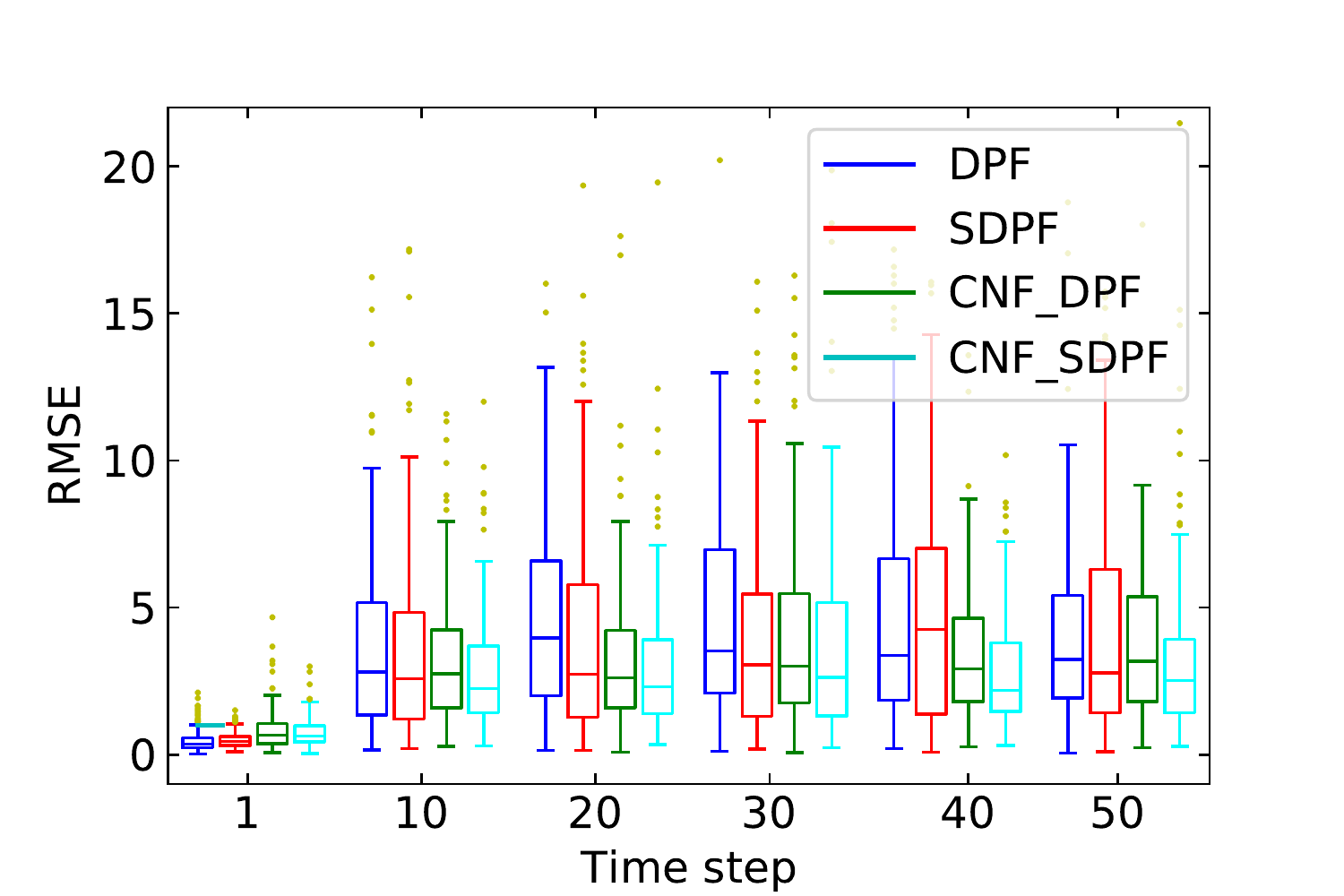}
    \caption{The boxplot of RMSEs computed with 150 testing trajectories at evaluated time steps (through three simulation runs).}
    \label{fig:boxplot}
    \end{center}
    \end{figure}

\begin{figure}[th]
    \begin{center}
    \includegraphics[width=.87\linewidth]{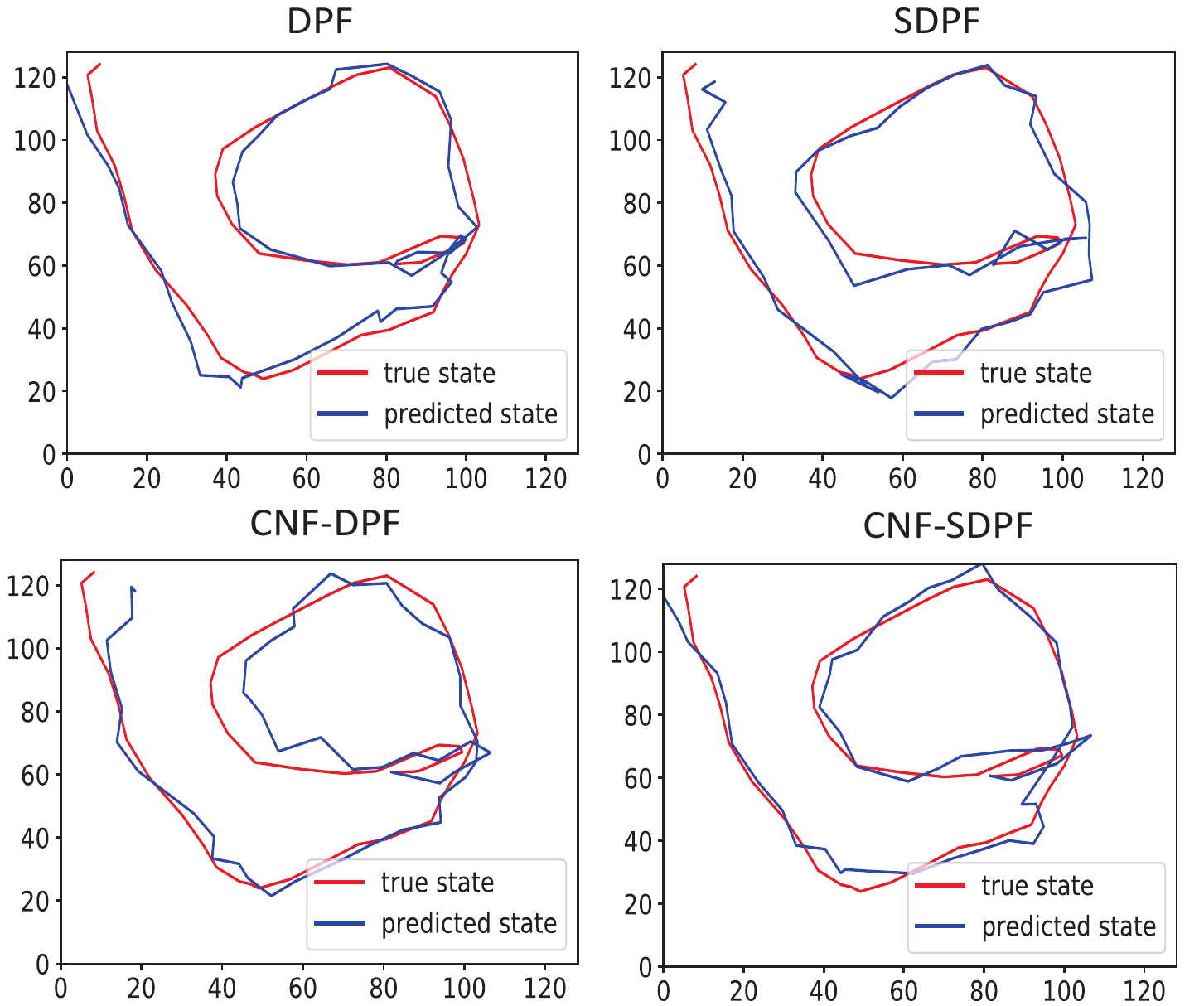}
    \caption{Visualization of the tracking results produced by the DPF, the SDPF, the CNF-DPF and the CNF-SDPF.}
    \label{fig:Tracking}
    \end{center}
    \end{figure}

Table \ref{tab:experiment1} shows the comparison of the tracking performance on the test set. The CNF-SDPF achieves the smallest prediction error, and for both the DPF and the SDPF, the incorporation of conditional normalizing flows can significantly improve the tracking performance in terms of the RMSE between predictions and true states. We observe that the conditional normalizing flow-based DPFs can move particles to regions closer to true states compared to their DPF counterparts, demonstrating
the benefit of utilizing information from observations in constructing proposal distributions. 
This is validated in Fig. \ref{fig:boxplot} which shows that the CNF-DPF and the CNF-SDPF consistently display smaller RMSEs at evaluated time steps with test data. Visualizations of the tracking results obtained by using different methods are presented in Fig. \ref{fig:Tracking}.
    
\begin{table}[t]
	\centering
	\caption{The comparison of RMSEs on the test set for DPFs, SDPFs, CNF-DPFs and CNF-SDPFs among three simulation runs (50 testing trajectories in each simulation)}.
	\begin{tabular}{l|l}
		\hline
		& RMSE  \\ \hline
		DPFs &  $5.479 \pm 0.105$ \\ \hline
		SDPFs & $5.544 \pm 0.058$ \\ \hline
		CNF-DPFs  & $4.539 \pm 0.066$ \\ \hline
		CNF-SDPFs  & $4.181\pm 0.105$ \\ \hline
		
	\end{tabular}
	\label{tab:experiment1}
\end{table}

\section{Conclusion}
\label{sec:conclusion}

We proposed differentiable particle filters (DPFs) which employ conditional normalizing flows to construct flexible proposal distributions. The normalizing flow is conditioned on the observed data to migrate particles to areas closer to the true posterior and we show that a tractable proposal distribution can be obtained. We demonstrated significantly improved tracking performance of the conditional normalizing flow-based DPFs compared to their vanilla DPF counterparts in a visual disk tracking environment.
\bibliography{ref.bib} 
\bibliographystyle{IEEEtran}

\end{document}